\documentclass[conference]{IEEEtran}
\IEEEoverridecommandlockouts
\usepackage{cite}
\usepackage{amsmath,amssymb,amsfonts}
\usepackage{algorithmic}
\usepackage{graphicx}
\usepackage{textcomp}
\usepackage{xcolor}
\usepackage{subfig}   

\def\BibTeX{{\rm B\kern-.05em{\sc i\kern-.025em b}\kern-.08em
    T\kern-.1667em\lower.7ex\hbox{E}\kern-.125emX}}
\graphicspath{ {images/} }

\begin{document}

\title{Towards autonomous ocean observing systems using Miniature Underwater Gliders with UAV deployment and recovery capabilities
\thanks{Research funded by the Research Council of Norway (RCN), the German Federal Ministry of Economic Affairs and Energy (BMWi) and the European Commission under the framework of the ERA-NET Cofund MarTERA, and by OsloMet lighthouse projects BiMUG and Autonomic Systems. NTNU-AMOS (Center for Autonomous Marine Operations and Systems) is funed by the Research Council of Norway through the Centers of Excellence funding scheme, Grant/Award Number: 223254-AMOS.}
}

\author{
\IEEEauthorblockN{Erik Sollesnes, Ole Martin Brokstad, Rolf Kl\ae boe, \\ Bendik V\aa gen, Alfredo Carella, and Alex Alcocer}
\IEEEauthorblockA{
\textit{Department of Mechanical, Electronics and} 
\\
\textit{Chemical Engineering. Oslo Metropolitan University}\\
Oslo, Norway \\
alex.alcocer@oslomet.no}
\and
\IEEEauthorblockN{Artur Piotr Zolich, Tor Arne Johansen}
\IEEEauthorblockA{
\textit{Centre of
Autonomous Marine Operations and Systems} \\
\textit{(NTNU AMOS), Department of Engineering Cybernetics}\\
\textit{Norwegian University of Science and Technology}\\
Trondheim, Norway\\
tor.arne.johansen@ntnu.no}
}

\maketitle

\begin{abstract}
This paper presents preliminary results towards the development of an autonomous ocean observing system using Miniature Underwater Gliders (MUGs) that can operate with the support of Unmanned Aerial Vehicles (UAVs) and Unmanned Surface Vessels (USVs) for deployment, recovery, battery charging, and communication relay. The system reduces human intervention to the minimum, revolutionizing the affordability of a broad range of surveillance and data collection operations. The MUGs are equipped with a small Variable Buoyancy System (VBS) composed of a gas filled piston and a linear actuator powered by brushless DC motor and a rechargable lithium ion battery in an oil filled flexible enclosure. By using a fully pressure tolerant electronic design the aim is to reduce the total complexity, weight, and cost of the overall system. A first prototype of the VBS was built and demonstrated in a small aquarium. The electronic components were tested in a pressure testing facility to a minimum of 20bar. Preliminary results are promising and future work will focus on system and weight optimization, UAV deployment/recovery strategies, as well as sea trials to an operating depth of 200m.
\end{abstract}


\section{Introduction}
\label{intro}
\begin{figure}
   \centering
   \includegraphics[width=0.98\linewidth]{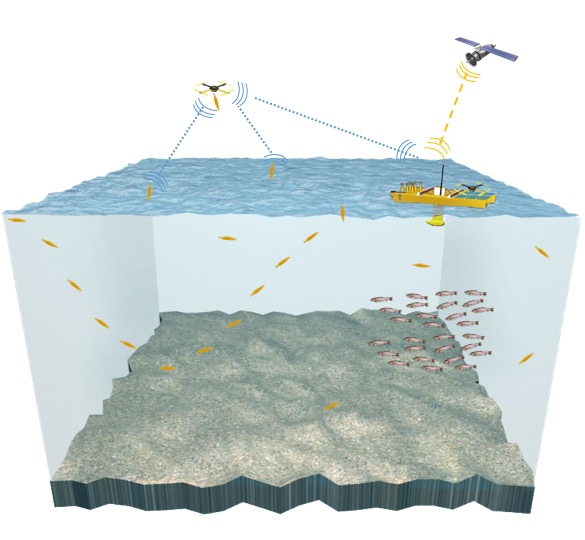}
    \caption{OASYS project concept 
    }
\label{fig:OASYSconcept}
\end{figure}

Autonomous robotic systems are steadily modernizing the way we obtain data and interact with the ocean \cite{ludvigsen2016network}. However, in the majority of the cases manned missions in the deployment/recovery phases are still required which represents a high percentage of the total operational costs. The OASYS project (Fig. \ref{fig:OASYSconcept}) will develop and demonstrate an innovative type of fully automated Ocean-Air coordinated robotic operation with the potential for drastically reducing the cost of ocean observing systems. The project proposes the development of a swarm of low cost Miniature Underwater Gliders (MUGs) that can operate autonomously with the support of Unmanned Aerial Vehicles (UAVs) and Unmanned Surface Vessels (USVs) for deployment, recovery, battery charging, and communication relay.

\begin{figure}[h]
   \centering
   \includegraphics[width=0.98\linewidth]{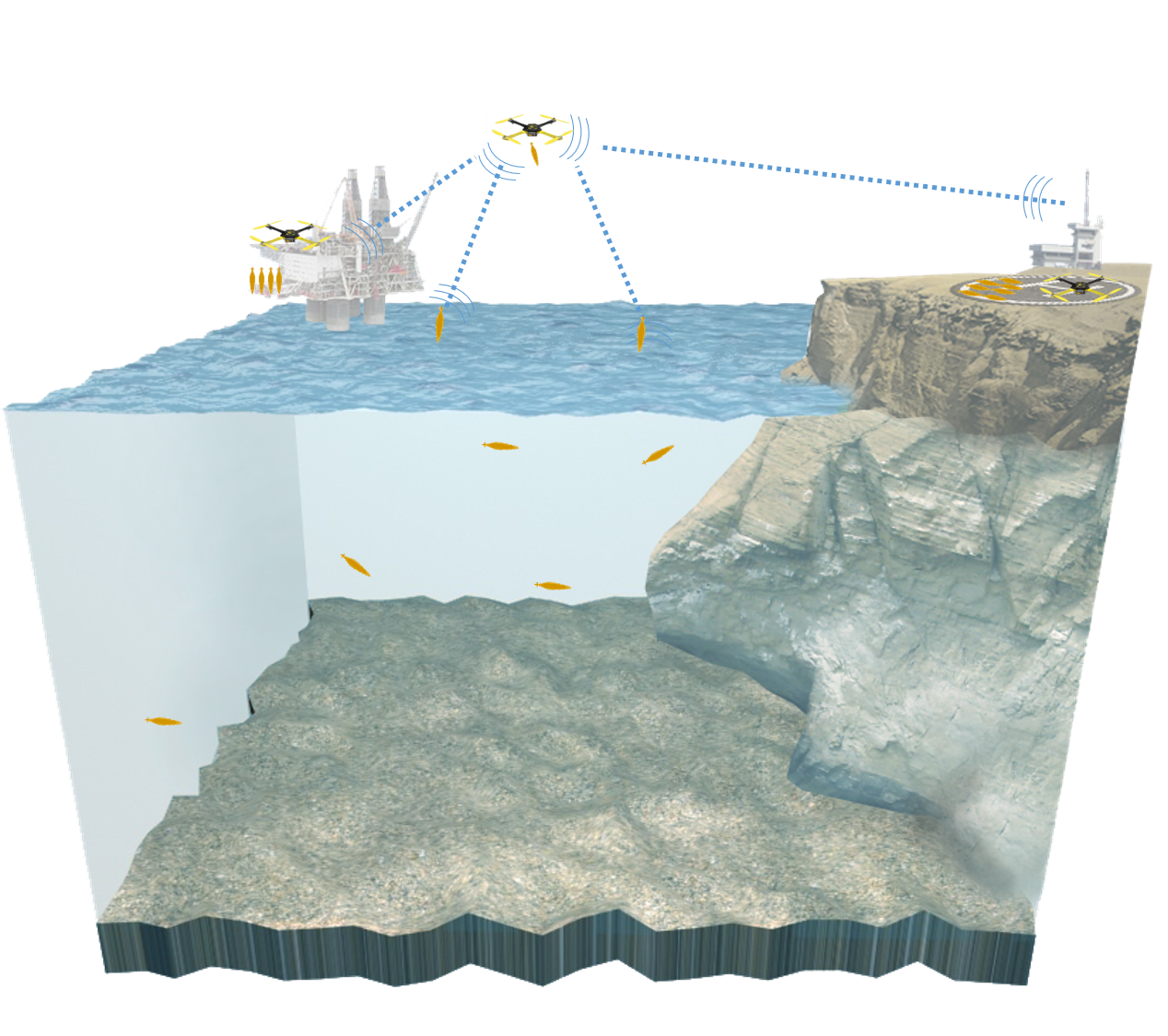}
    \caption{OASYS project concept of operation in coastal areas or in the vicinity of offshore structures.}
\label{fig:scnario2}
\end{figure}

When operating in coastal areas or in the vicinity of offshore structures, multirotor type UAVs can be used for deployment and recovery while fixed base stations are used for battery charging, see Fig. \ref{fig:scnario2}. For long endurance operations, USVs can be used as motherships for both MUGs and UAVs, and serve as battery charging station and satellite communication link.

The propposed MUGs have severely limited characteristics as compared to existing commercial underwater gliders. In order to allow UAV deployment and recovery the weight has to be reduced drastically which imposes limitations on the payload and battery capacity. Battery capacity allows operation for a few days, as compared to months. Communication capabilities are also reduced. An RF link allows to communicate with UAVs while in line of sight  \cite{zolich2016communication}, but unlike their larger siblings, MUGs are not equipped with satellite communication capabilities. Also operational depth will be initially limited to 200m, although future designs will aim at higher depths. 

This paper presents preliminary results towards the development of MUGs and specifically a core component which is their Variable Buoyancy System (VBS). Variable buoyancy systems are a fundamental component of underwater gliders, drifters, and some autonomous underwater vehicles as they allow to control vertical speed and in the case of gliders forward motion \cite{rudnick2004underwater} \cite{MacLeodIEEEJOE2017}
\cite{sahoo2016computing} \cite{waldmann2016performance}. There are several types of VBS including pumped water, pumped oil, piston, and thermal expansion driven \cite{jensen2009variable} \cite{webb2001slocum} \cite{Shibuya2009DevelopmentOA} \cite{yamamoto2017new}. Recently there has been increasingly interest in the development of miniaturized VBS to be used in underwater vehicles \cite{yamamoto2017new} \cite{zhang2014miniature} and drifters \cite{jaffe2017swarm}.

\section{Miniature Underwater Glider (MUG) development}
\subsection{Variable Buoyancy System concept}
\label{section2}
This paper proposes the development of MUGs with a miniature VBS in a fully pressure tolerant electric design. The goal is to reduce the cost, size, and overall complexity, by eliminating the need for external pressure housings. The VBS proposed is based on a small piston driven by a brushless DC motor. The piston is enclosed in an oil filled body with a flexible bladder. Changing the piston position displaces up to 100cc oil, which expands the bladder and changes the overall system volume.

\begin{figure}
   \centering
   \includegraphics[width=0.98\linewidth]{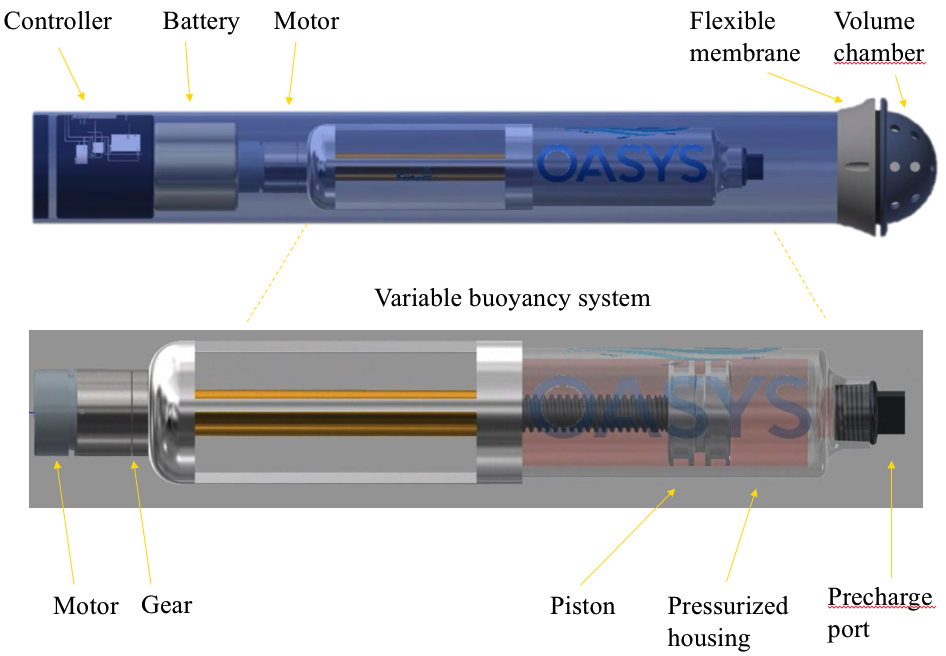}
    \caption{Pressure tolerant Variable Buoyancy System preliminary prototype}
    \label{fig:MUGdiagram}
\end{figure}

All components are immersed in oil, and the only pressure housing is the cylinder containing the piston, see Fig. \ref{fig:MUGdiagram}. By adopting this design there is no need for an external pressure housing. All components can be enclosed in an inexpensive flexible container, or bladder, which in turn can help reduce the total volume, and weight of the system. This is of critical importance, as the end goal of the project is to minimize weight (in air) such as to enable deployment and recovery using Unmanned Aerial Vehicles (UAVs).
The prototype is initially designed for an operational depth of 200m. Future designs will aim at higher depths. 
\subsection{Motor and drive}
The motor used to position the linear actuator is a Maxon brushless DC motor with integrated hall sensors. The driver is a Ingenia Neptune driver using the hall sensor feedback for position control.

\begin{figure}[h]
   \centering
 \subfloat[motor in air, ambient pressure]{\includegraphics[width=0.98\linewidth]{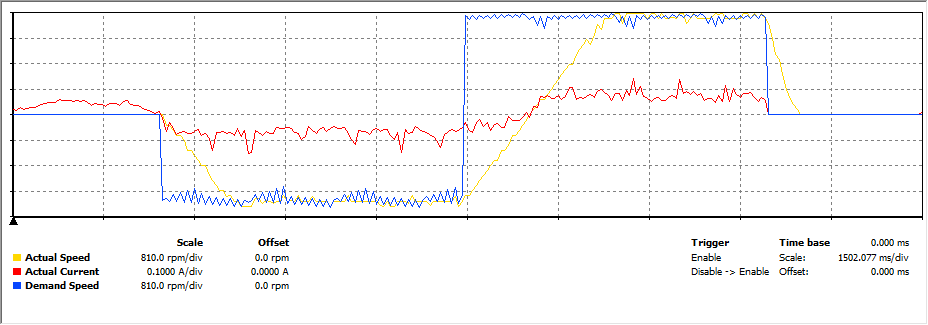}}\\
\subfloat[motor in oil, ambient pressure]{\includegraphics[width=0.98\linewidth]{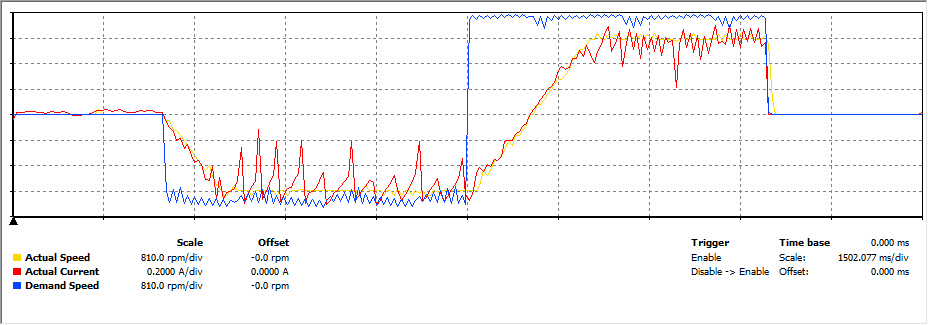}}\\
\subfloat[motor in oil, 55bar]{\includegraphics[width=0.98\linewidth]{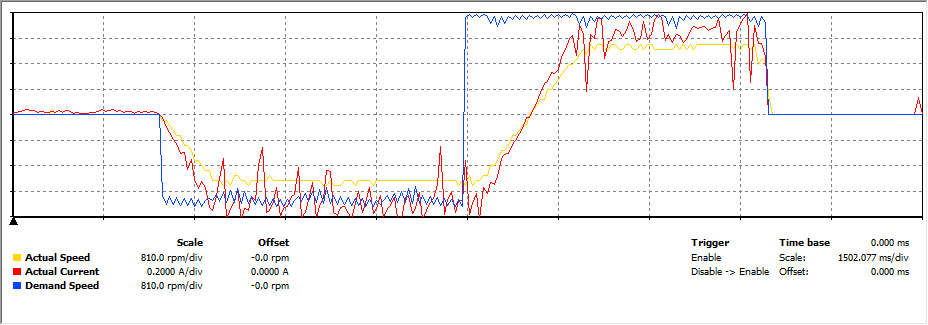}}
\caption{Motor testing showing setpoint speed (blue), actual speed (yellow), and current (red) in different operating conditions.}
\label{fig:motor_graphs}
\end{figure}

The motor was tested in air, immersed in oil, and immersed in oil under 55bar external pressure as shown in Fig. \ref{fig:motor_graphs}.

\subsection{Battery}
\begin{figure}[h]
   \centering
   \includegraphics[width=0.98\linewidth]{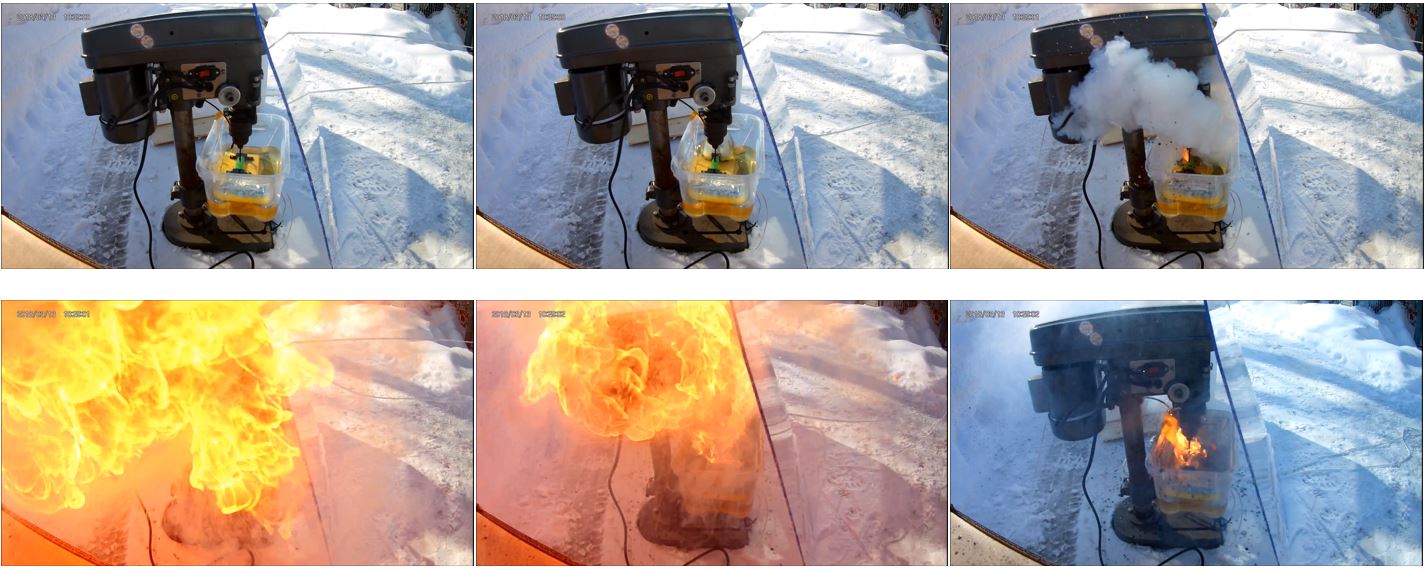}
    \caption{Battery nail penetration destructive test.}
    \label{fig:video_battery_nailtest}
\end{figure}

Powering the VBS is a battery pack (consisting) of six (lithium ion) LG MJ1 cell batteries connected in series. With a standard charge this gives a voltage of 25.2V and a capacity of 3500mAh. A destructive nail penetration test was performed in one individual (over-charged) cell, while immersed in dielectric oil Fig.\ref{fig:video_battery_nailtest}. The battery pack was also pressure tested under operating conditions suggesting that the lithium ion battery pack can survive pressures of at least 20 bar. 
\begin{figure}[h]
   \centering
   \includegraphics[width=0.55\linewidth]{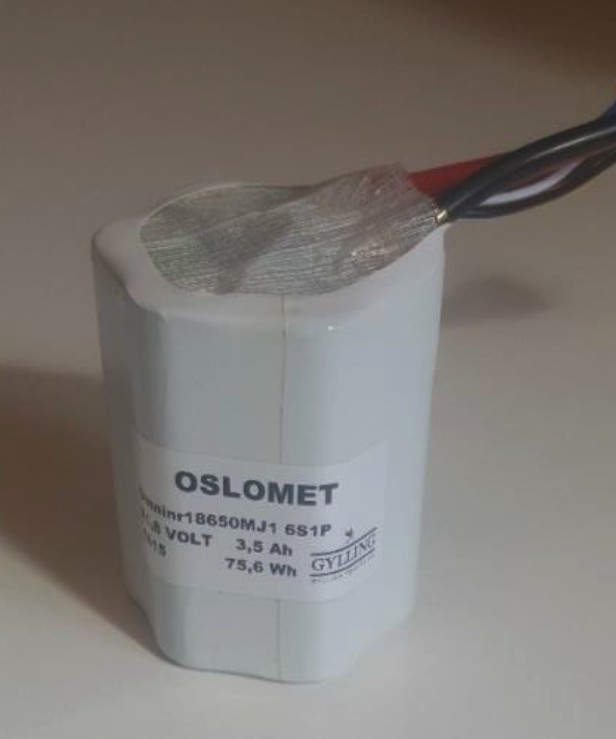}
    \caption{Battery pack, lithium ion 6S 3500mah.}
    \label{fig:batterypack}
\end{figure}
\subsection{Microcontroller and RF module}
The microcontroller used for controlling the VBS was a Moteino MEGA made by LowPowerLab. It's an Arduino based microcontroller with built in RF capabilities, and it's designed with low power consumption in mind. The Moteino MEGA was operated at 433MHz frequency, and the antenna was 173mm long copper wire.
\subsection{Payload}
Due to the reduced size of MUGs, payload capabilities are limited, and a compromise on the type of sensors and accuracies will need to be made.  Current design includes a miniature DST CTD from Star-Oddi which has a weight of 21g, and has an option for external power supply and real time data transmission. The MUG prototype is also equipped with a Keller PA7LC pressure transducer. The goal is to incorporate a miniature CTD, and miniature fluorometer developed by TriOS gmbh, which is one of the partners in OASYS project. The fluorometer can be targeted to detect concentration of a large number of biological and chemical parameters including hydrocarbons, algae pigments, and dissolved oxygen.
\subsection{Navigation Guidance and Control systems}
Due to their reduced size and cost, navigation capabilities are also severely reduced. MUGs are equipped with a GPS receiver, which allows for position fixes while on the surface, but navigation accuracy underwater is limited and probably limited to dead reckoning. A miniature low cost acoustic modem which allows for short range communication of simple commands, and maybe positioning information, as inspired by \cite{katzschmann2018exploration}, will be evaluated. 

\section{UAV and USV support}
The MUG is designed to be carried by a long-endurance USV to its deployment spot. Available USVs, e.g. AutoNaut (AutoNaut Ltd., UK), offers sufficient cargo volume and capabilities to perform that task.

One of the important aspects of long-endurance USVs operations is power management. Although vehicles is driven directly by waves, and uses virtually no energy, control electronics requires constant supply of power. Vehicles are then equipped with batteries and recharging system that harvest, e.g. solar energy.

\begin{figure}
   \centering
   \includegraphics[width=0.98\linewidth]{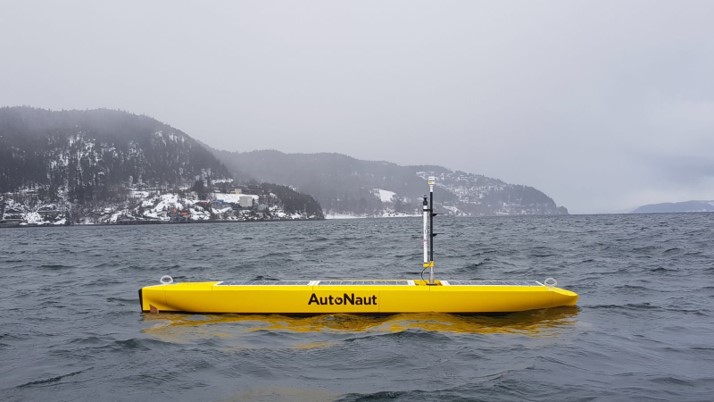}
    \caption{AutoNaut long range wave powered USV owned by NTNU}
    \label{fig:autonaut}
\end{figure}

In addition to MUG, the USV will accommodate also multirotor UAV, and support equipment that will allow data exchange between units and energy transfer. The UAV, prepared to operate in marine environment, e.g. HexH2O (XtremeVision360 Ltd, UK), will carry two main roles. First, it will be used to collect the MUG from the surface, after it has finished its mission, and deliver it back to the USV. A certain highlight of the UAV use is significant energy consumption required for flight, which conflict with limited energy available on-board the USV. Recharge of the UAV battery has to be carefully managed, and flights adjusted appropriately to the forecast energy levels.

Second role of the UAV can be a radio signal relay between MUG and the USV. A reliable direct data-link between MUG and USV may be a challenging task. Due to low elevation of the MUG's antenna above the water level direct line-of-sight between MUG's and USV's transceivers may be intermittent. For that reasons, UAV can be utilized as an elevated antenna, providing a relay-link between MUG and USV. The communication technology between vehicles will base on modern, self-configurable IoT solutions, which support dynamic and scaleable networks.

The UAV can also support MUG operations during missions where USV is not used, e.g. for deployment and recovery from a support vessel, or in a coastal area, see Fig.\ref{fig:scnario2}.
Recovery of MUG directly from a ship may be challenging due to limited maneuverability of MUG and the ship, and may otherwise require use of a small manned boat. High maneuverability of the multirotor should allow recovery of the MUG in a safe distance from the ship, away from the turbulence created by its hull and propellers. Similarly, during deployment or recovery in a coastal area, UAV can deploy or pick up a MUG in a safe distance from obstacles and where the depth is sufficient.

Data from the MUG collected at the USV can be then forwarded using main data-link to the user. Users of the system will be able to monitor and control operations of the vehicle using a Situation Awareness (SA) system, e.g. LSTS Toolchain~\cite{6608148}.
The system will provide a information about last known positions of the vehicles, and their tasks. It should also allow data visualization in the mission context.

\section{Preliminary results}
\begin{figure}
   \centering
   \includegraphics[width=0.98\linewidth]{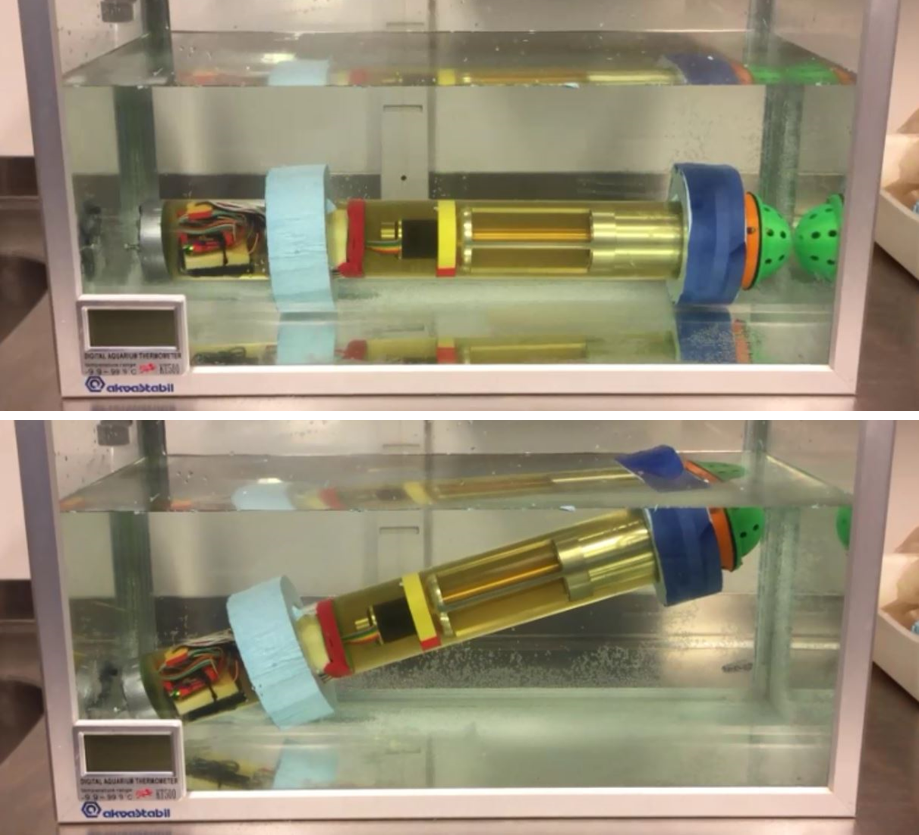}
    \caption{Prototype VBS functionality demonstration in a small aquarium. Total weight of 2,6kg diameter of 7cm and length of 56cm.}
    \label{fig:MUG_demo}
\end{figure}

All components used in the preliminary prototype shown in Fig.\ref{fig:MUG_demo} were tested under at least 20 bar pressure in oil. This includes the brushless DC motor, the Li-Ion Battery pack, the Moteino microcontroller and the Neptune driver. The brushless DC motor pressure test results shows the motor drawing about five times the current in oil compared to air. The motor is not reaching set point speed due to a maximum continuous current rating of 0.5A. The results show that the motor performs similar in oil and at ambient pressure compared to 55 bar pressure. The Moteino RF communication was successfully tested through a small amount of oil and water, as shown in Fig.\ref{fig:MUG_demo}. The prototype tested in a small aquarium, shown in Fig.\ref{fig:MUG_demo}, demonstrates the VBS basic operation.
\section{Conclusions and future work}
The prototype VBS described in this paper demonstrates the viability of the OASYS MUG concept. The prototype is able to change its buoyancy with the VBS enabling vertical motion. The individual components are shown to be pressure tolerant retaining functionality at  pressure equivalent to 200m depth. This early prototype serves as a foundation upon which a completed MUG can be developed. Further development will add the ability to adjust pitch and yaw, improve power efficiency, add GPS and environmental sensors, as well as UAV deployment/recovery strategies.

\section{Acknowledgements}
The authors would like to thank Eirik Fossdal, Erik Fidje, Jenny Tran, Ole Jacob Brunstad, and Rune Orderl\o kken for their help in the development of the mechanical prototype, and Gylling Teknikk AS for their help in the development of the battery pack. 
\bibliographystyle{IEEEtran}
\bibliography{main}

\end{document}